# The meaning of manufacturing know-how

*Kryssanov[1], V.V., Abramov[2], V.A., Fukuda[3], Y. and Konishi[4], K.*
[1] *Kobe University, Kobe, Japan;*
[2] *IACP of FEB RAS, Vladivostok, Russia;*
[3] *Hosei University, Tokyo, Japan;*
[4] *TRI of JSPMI, Tokyo, Japan*
*E-mail: kryssanov@ziong.cs.kobe-u.ac.jp*

**Abstract**
This paper investigates the phenomenon of manufacturing know-how. First, the abstract notion of knowledge is discussed, and a terminological basis is introduced to treat know-how as a kind of knowledge. Next, a brief survey of the recently reported works dealt with manufacturing know-how is presented, and an explicit definition of know-how is formulated. Finally, the problem of utilizing know-how with knowledge-based systems is analyzed, and some ideas useful for its solving are given.

## 1 INTRODUCTION

In the late 90th, the complex of concepts, theories, technologies and software called knowledge-based systems has become a key point in development of many future-oriented manufacturing paradigms, such as Agile Manufacturing and Intelligent Manufacturing Systems. Besides, the progress from craft production, to automated and flexible production and towards 'next generation' production is now realized to be in many respects determined by the human/systems ability to handle the domain knowledge rather than simply by a given standard of knowledge in the domain. This is the motivation for a continuously growing research interest to utilization of manufacturing knowledge.

It should be noted however, that while a great many reports on different theoretical and applied aspects of knowledge utilization have been published, the issues of the specificity of manufacturing knowledge and the appropriateness of the methodologies brought into manufacturing from other domains to build knowledge-based systems have not been given due attention. One instance of this research lack is given in this paper with the phenomenon of know-how.

It was discovered rather long ago that know-how plays an important role during the solving of professional tasks in manufacturing (e.g. Iwata and Fukuda, 1986). Furthermore, it is often claimed in the specific literature and mass media that a success of a particular manufacturer is to a large extent related to the possession of some know-how by this manufacturer. It would then be logical to presume that utilization of know-how should significantly take place in current manufacturing research, resulting in various 'know-how -based' application systems. However, the authors' experience of dealing with know-how while developing knowledge-based systems (Kryssanov et al., 1997) has been that the phenomenon of know-how is still unexplored and poorly grasped by the manufacturing research community. What is worse, there is no explicit definition of know-how nor evident comprehension of what is behind this term.

The objective of this work is to scientifically interpret manufacturing know-how and discuss feasible approaches to utilizing know-how with computer systems.

## 2  KNOWLEDGE

By Newell, knowledge is what an observer ascribes to an intelligent agent that allows the observer to construe the agent's behavior as rational (Newell, 1982). Such a view of knowledge is general enough to impose no restrictions on the knowledge form, content or verity. If knowledge is expressed or represented in an explicit form, it can be treated as information. Conversely, information becomes knowledge when it is purposefully and sensibly applied by an agent. Below, an attempt is made to highlight a set of the most common distinctions about knowledge to set the base for defining know-how.

One of the principal distinctions, which appear while classifying knowledge, is the differentiation of scientific and technological knowledge. A major feature that distinguishes them is purpose or intention (Herschbach, 1995). The purpose of scientific knowledge is to understand and explain phenomena and the laws of nature. Science does so trying to investigate separate views of reality. Scientific knowledge is about analysis and knowing. Contrasting this, the purpose of technological knowledge is praxiological, i.e. to manipulate objects in the physical world, to create artifacts. Technological knowledge is about synthesis and efficient application.

The difference between scientific knowledge and technological knowledge is not antagonistic in manufacturing: since the synthesis process requisite to create a complex manufactured article can hardly be performed without analyzing all the aspects associated with the article development, modern production technologies necessarily include a large amount of scientific knowledge. Technological knowledge makes use of scientific knowledge, and they both have to be employed for solving the real-world problems. However, the role of the former in manufacturing task solving is primary, and scientific manufacturing knowledge is usually dealt with by embodying it into a technological context.

Frey (1989) introduced four general levels of technological knowledge. The first level is to contain craft skills that are usually implicit knowledge taught through direct observation, imitation or trial and error rather than through discourse. Technical maxims and 'rules of thumb' constitute the second level. They are obtained by summarizing an experience, and they are expressed explicitly but are still applicable and understandable only in close conjunction with the on-going activity underlying these rules. Empirical laws, that establish the next level, are formalized generalizations derived from experiences. They are not yet scientific because such laws do not make claims about their generality or wideness of applicability, but are intended to support particular task solving processes. Technological theories are at the fourth level. They systematically use scientific knowledge and provide a coherent framework for the professional activities.

It can be seen that all the levels are severed by the context-dependency of the represented knowledge. This knowledge is highly situated, contingent, local and environmentally dependent on the first ('wet') level, but it is relatively generic, abstract, and to some extent context-insensitive on the fourth ('dry') level. It is important to note that these levels are complementary, and none of them alone are sufficient to adequately grasp the real-world problems: there are many interconnections among the levels, formed in different ways at different scale abstractions (see Goguen, 1992).

At any of the levels, the knowledge can be explicit and implicit as well. The distinction between explicit and implicit knowledge is based on the comparative easiness of knowledge acquisition, sharing and use. If knowledge is readily expressed in a comprehensible form (e.g. presented by an agent in a well-formalized form) and can be utilized immediately, then the observer, who is to use this knowledge, could call it explicit. At the same time, another knowledge can be understood only with a substantial mental effort (e.g. through reasoning) or may not be received at all. Such knowledge could be considered as implicit. Unconscious abilities, skill, ingenuity, and the like are of the knowledge kind that is mainly implicit from the observer viewpoint (though usually for the agent also). This kind of knowledge is extremely difficult to reproduce and/or represent. Due to the specificity of some kinds of implicit knowledge, apparently conditioned by the nature of human knowledge representation, such 'hidden' and largely inexpressible knowledge has been called tacit (Polanyi, 1966).

Another important distinction about knowledge kinds that is regularly seen in the literature is the classifying of knowledge into descriptive and prescriptive, or factual and procedural respectively (Berry, 1995; Vincenti, 1984). This corresponds to the separation of knowledge into the knowledge about the environment where an agent acts and the knowledge about what has to be done by the agent in order to achieve the desired result. Descriptive knowledge is associated with the form 'know that' of natural language, while prescriptive knowledge matches the form 'know how' (that is not know-how).

Berry (1995) added the third knowledge category – 'know why' and called it judgmental knowledge. He put goals and constraints into this category. It seems to be reasonable because 'know that' and 'know how' appear as an inalienable part of task solving whereas the task itself, its stipulations, scope and specificity remain ambiguous. Why a deed is to be done, why it should be done in this way, why this way is better, etc., – 'know why' knowledge is necessary to answer such questions. However, these questions have to be addressed in the global context of the human activities rather than within the limited context of a task. Therefore, the knowledge 'know why' is to specify the global-level epistemological validity of particular pieces of descriptive and prescriptive knowledge used to solve tasks.

Considering all of the above distinctions, one may notice that there are no strict dichotomies but something like intersected (although not coincident) areas, which set up the dimensions or continua of knowledge, where each of these dimensions – scientific-technological, 'wet'-'dry', explicit-implicit, descriptive-prescriptive-judgmental – fits to a certain phenomenon or aspect of the use of knowledge.

## 3    MANUFACTURING KNOW-HOW

The Concise Oxford Dictionary defines know-how as 'practical knowledge; technique, expertise...' that is a rather wide interpretation and definitely not satisfactory for our purposes. More detailed, though situated, context-specific explications of know-how have been found in the literature as reviewed below.

A knowledge-based diagnosis system was developed to utilize 'comprehensive know-how about maintenance and the equipment' (Wincheringer and Miklavec, 1996). That know-how includes 'knowledge about the assembly and function of the machinery, the course of production processes and information about which measures are to be carried out at which time.' The authors emphasized that the know-how about maintenance can be acquired by way of instructions and training only. The latter commonly implies an explicit and prescriptive character of the underlying knowledge. On the other hand, the know-how about the equipment was treated in the study as descriptive knowledge.

An expert system qualified to arrange the polishing process using 'experienced know-how' was described in Saito et al. (1993). The knowledge adopted into the system is the results of experiments on the polishing performance of whetstone, which have been formalized in the form of empirical laws. Such knowledge could be classified as descriptive and doubtless explicit.

In Ismail et al. (1995), that is a case study on the development of an intelligent system for progressive die design and manufacturing, the authors noticed that it is very difficult to elicit (and hence to share) 'the design know-how from experienced designers.' It can be inferred from the research that this difficulty is caused by the tacit peculiarity of that type of know-how.

A decision-making support system based on know-how was presented in Kryssanov et al. (1997). The system has been designed to support process planning, and its knowledge base contains the information extracted from a representative collection of practical know-how dealt with for end milling. Know-how compiled in the collection contains descriptive and prescriptive knowledge represented as empirical laws.

Chep and Anselmetti (1993) suggested a conceptual model for the representation and use of manufacturing knowledge. The model was intended for the automation of process planning. It was argued that manufacturing know-how is subjective and volatile and usually requires analysis to be represented in a readable form. The knowledge of know-how was characterized as implicit and context-dependent.

The work of Mahé et al. (1996), dedicated to the problem of know-how capitalization in an enterprise, made another commitment regarding the notion of know-how. It was suggested distinguishing between two types of tacit knowledge: 'context knowledge' that is 'a set of norms and implicit values more or less shared (culture, behavior, ...)' and 'practice knowledge (know-how)' a 'particular expertise that allows (one) to realize something easily and efficiently and that is acquired by experience.' A definition of know-how as 'a set of complex and not easily to formalize knowledge which human beings have acquired through a process of learning, reasoning with analogies and intuition' can be found in Sellini and Vargas (1996). The idiosyncratic, peculiar (if not private) and tacit character of know-how was similarly pointed to in research less centered on the problem of know-how utilization (Kyläheiko, 1997; Choo, 1995).

Considering the above, we can conclude that any type of knowledge can play the role of know-how, and the surveyed classification of knowledge dimensions at this time is not complete enough to firmly define the phenomenon of know-how.

Wincheringer and Miklavec (1996) declared that their knowledge-based diagnosis system could be used for 'safeguarding the diagnostic know-how of a particular company.' In Schulz and Spath (1997), know-how was considered as knowledge that is possessed by an individual qualified operator but that may not be known to designers and planners. Chep and Anselmetti (1993) asserted that knowledge underlying know-how is shared, but is partial and personal. These examples and many others similar, but not cited here, as well as our own recent experience in dealing with knowledge classified as know-how (Kryssanov et al., 1997), have led us to the necessity of using the extension of traditional knowledge classification to cover the know-how 'dimension.'

It should be noted first, that in the literature and mass media, the term know-how bears the stamp of an information commodity that just provides its possessor with a business success advantage over the competitors or, in other words, gives one the possibility to solve professional tasks more efficiently than others do it. The study described in Cainarca et al. (1997) strengthens this supposition. The authors in this work discussed know-how as one of the major factors, which create a competitive advantage and give a company desired sustainability. Similar ideas can be found in Pirskanen (1997), where a software toolbox designed to increase quality and competitiveness of small and middle-sized companies in Finland was outlined.

We will argue that to obtain a comprehensive image of know-how, the entire bulk of the domain knowledge, its distribution, transfer and evolution should be considered. The following three 'axes' (dimensions) of knowledge classification are necessary to define know-how: spatial, temporal and conceptual.

The distribution of knowledge among the agents (e.g. companies, people, computers) can be correlated to the spatial dimension. For example, while one company owns some knowledge (or information), another does not though may need this knowledge. In this case, there are two distinct representatives being potentially capable of possessing specific knowledge. Generalizing, the spatial dimension is to allow for the existence of many individual agents in the domain.

The temporal dimension is responsible for the evolution of the knowledge of the agents over time. New materials, tools, machine tools, etc. continuously appear, and information about their properties, capacities and application proliferates. Also, new methods and approaches to task solving, successful (and otherwise) experiences are generated, accumulated and disseminated. Commonly, new knowledge supplements and eventually replaces old knowledge to fit the new requirements of the social and technical environment. The temporal dimension is to take into account the phenomena of relative renewal as well as the obsolescence of the domain knowledge.

The conceptual dimension concerns the conceptual pertinency of knowledge. In fact, the domain knowledge can naturally be ordered from common to the whole domain, generic knowledge to particular, even specific knowledge relevant to the practice of an individual agent. The conceptual dimension is to arrange the domain knowledge by professional activities.

We introduce the following definition of know-how.

*Given two agents* (the spatial dimension), *a piece of the first agent's technological knowledge can play the role of know-how, considering it from the second agent's viewpoint, if this piece is on the potentially shareable locality level of both agents, i.e. it is relevant to the tasks and the tasks solving of the agents* (the conceptual dimension), *and is an innovation in respect to the knowledge of the second agent* (the temporal dimension).

Thus, we want to stress that there is no special single kind of knowledge behind know-how but that it is just a role that can be filled by technological knowledge (as well as by scientific knowledge embedded into a technological context). Hence, no knowledge of a single agent can be considered as know-how unless in conjunction with an observer (another agent) and in the presence of the possibility of knowledge sharing. As a corollary, when building knowledge-based systems, which are intended to utilize know-how, one should not try to find an extraordinary means to capture and represent knowledge underlying know-how but should look for an effective mechanism to update the systems knowledge.

# 4  FROM KNOWLEDGE SHARING TO KNOW-HOW UTILIZATION

Having clarified the meaning of manufacturing know-how, we will argue that handling know-how with knowledge-based systems demands introducing a special functionality for these systems. This functionality is in essence the systems ability to mutually communicate and share knowledge.

There are several preconditions required to enable knowledge sharing among the agents. First, the agents should have appropriate information transmitting facilities. This is a necessary but not sufficient condition because 'raw' information (e.g. data), being exchanged, gives no knowledge without interpretation of this information. The agents need a means of knowledge (information) representation, which could allow them not only to represent each agent's knowledge, but also to attach the meaning of the information content to the resulted representation. However, such a meaning attachment and the further use of one agent's knowledge by another agent for task solving can have an arbitrary relevance to the tasks of the second agent if there are no facilities to properly 'synchronize' the interpretation of the agents' knowledge along the temporal and conceptual dimensions. The agents should then have a 'common ground' for knowledge interpretation instead of relying on 'by default' concealed assumptions.

For a long time, knowledge representation as a field of AI has been mainly concerned with the study and development of various formalisms and artificial languages. The problem of knowledge sharing has been dealt with by assigning a rigid representation layer for all the agents who participated in the exchange. Common vocabularies, lexicons, taxonomies, ontologies, inference syntaxes, tasks, application protocols, and problem-solving methods have often been discussed as candidates for the common ground of knowledge sharing, and a number of studies on the subject have been published.

Guarino (1994) described a fundamental classification of knowledge representation systems. This classification placed a knowledge representation formalism onto one of five levels: logical, epistemological, ontological, conceptual, or linguistic. A representation formalism at the logical level offers such primitives as relationships, predicates and functions, which are usually to have the standard formal semantics. This level is to assure logically correct formalization. The epistemological level is to offer knowledge structuring primitives such as structural constant, role, class, etc., as well as interrelationships among them (e.g. causality, generalization, hierarchy), not introducing any new semantics different from that which were fixed at the logical level. The ontological level serves to arrange an explicit specification of ontological commitments, which introduce the meaning for knowledge structuring primitives of the previous levels. This specification restricts the number of possible interpretations of the knowledge expressed in a formalism by tying the formalism's constructions to the patterns and regularities of the objective reality. The conceptual level is to handle particular instances of domain concepts. Due to the cognitive nature of these instances, they have only partially formalized semantics that are formed by inheriting the semantics of the previous levels. Finally, the linguistic level delivers nouns, verbs and other parts that are directly used by people to represent knowledge.

Taking this classification as a foothold and reviewing the literature, we have distinguished three primary groups of knowledge sharing approaches: 'logical', 'ontological' and 'common sense' -based.

The first group – logical – is mostly associated with Gruber's work on a translation approach to knowledge sharing (Gruber, 1993). The essence of his idea is to provide the agents, which need to share knowledge, with a means to translate the agents' knowledge among epistemological-level formalisms. This can be done through a comprehensive logical level of a translation mechanism developed in the study. In fact, the author's approach leads to the establishment of a common logical layer for all the communicating agents. Knowledge exchange is implied to take place after making preliminary agreements about the context of the intended use of the knowledge being of the participating agents' mutual interest. There is no reference model provided but only local informal regulations stipulated to co-ordinate interpretation of the knowledge. Gruber's translation approach makes the logical level the only computationally enforceable representation ground for the agents, leaving the context of knowledge use weakly (if at all) formalized. An immediate consequence of such a design decision is that only a part of the domain knowledge, which is completely formal and relatively 'dry' by its nature, can be shared successfully, i.e. it can correctly and uniquely be interpreted by all the agents. However, utilization of more complex kinds of knowledge will crucially depend on the informal preliminary co-ordination among the agents. The latter is a very subjective and scarcely controlled process.

Another idea was described in Guarino (1995), where the necessity of unambiguous and valid specification of the ontological level in a knowledge representation formalism was accented as the key point to reuse knowledge written in this formalism. Ontology in general is a branch of philosophy dealing with the *a priori* structure of the reality. Then, it seems rational to strive for explicit specification at the ontological level, where knowledge structuring primitives should obtain their association with the reality – the most truthful interpretation principally possible under given conditions. However, the need of the ontological level specification imposes some extra requirements on the used formal system: the possibility to adopt a supplementary, more expressive language to specify the ontological level commitments or the development of the appropriate syntactic facilities within the same formalism to keep ontologically meaningful distinctions among its primitives.

The ontological approach would in principle provide a reliable common ground for the knowledge exchange. However, it is not always possible nor necessary to share knowledge by way of the common ontological level. For example, many of the existing abstractions in science (e.g. in physics) as well as a majority of the social theories can hardly be related directly with the reality, and therefore they may not be dealt with by Guarino's approach.

The last but not the least, the third 'common sense' group is formed by research, in which the conceptual and linguistic levels have been chosen to synchronize interpretation of shared knowledge. A recent example of this group's approach is the method of 'a broad coverage ontology' described in Swartout et al. (1996). The authors suggested a large conceptual lattice including as many terms of the natural (English) language as possible to support information (knowledge) models creation and knowledge sharing as well. Each concept in the lattice corresponds to a word sense, while all the concepts are ordered by subsumption relations. A fragment of the lattice can be taken to relevantly be adapted to a particular context of the concept usage. Then, the lattice serves as a global reference model, and the agents exchange their knowledge, involving common sense concepts of the lattice.

The gist of the broad coverage ontology approach is that concepts underlying natural language are to play the role of the common ground for sharing professional knowledge. It can be postulated that possessing human natural language implies possessing 'common sense' knowledge. Methodologies of the third group sustain knowledge sharing, provided such 'common sense' unified knowledge exists.

Considering the fairly different groups of knowledge sharing techniques above, one may notice that each of them is innately oriented to exchange specific kinds of knowledge. The literature reveals many examples of techniques and tools for knowledge sharing, which tend to be rather heterogeneous in respect to the groups – different parts of knowledge are tackled by different approaches. Unfortunately, none of the contemporary knowledge sharing paradigms alone offers a theory sufficiently complete to cope with the exchange of manifold know-how.

## 5   SOME SPECULATIONS AND CONCLUSIONS

In the previous sections, a few of the important issues concerned the study of knowledge, knowledge representation and knowledge sharing have been examined with respect to the phenomenon of manufacturing know-how. It is understood that much has still to be done to pragmatically address the problem of utilizing know-how with knowledge-based systems. Derived from the authors' experiences, starting points for the development of 'know-how -based' manufacturing applications would be as follows.

The spatial and temporal dimensions of know-how could be addressed by establishing networked repositories (funds) of manufacturing knowledge and defining standard procedures/protocols for accessing the repositories. Today, the latter is the objective of many ongoing academic and industrial projects (e.g. van Heijst et al., 1996; Regli and Gaines, 1997). Under the assumption that, to be exchanged and further utilized, know-how must be expressed in an explicit and decipherable form, dealing with the conceptual dimension of know-how requires the development of appropriate translation mechanisms, which are to establish coordinated links between different knowledge representation systems practiced in the domain. Such links can be set up in three distinct ways: 1) fixing a rigid mapping between two notations at the logical or/and epistemological level(s); 2) assigning a mapping between two notations at the logical or/and epistemological level(s), which is to be regulated according to formal descriptions of ontological commitments of knowledge structuring primitives; 3) or via a language-independent conceptual lattice, involving (when necessary) information from the ontological levels.

We adapt the definition of the term 'ontology' as 'an explicit, partial account of a conceptualization' (Guarino and Giaretta, 1995), interpreting a conceptualization as a rational mental process of analyzing the reality by an individual agent or by a socially (environmentally, educationally, culturally, etc.) uniform group, resulting in an abstract, formalized, and simplified view of the world that is to be represented. Then, provided that instances of know-how representations are properly supplemented with ontological descriptions, the third of the above translation schemes suggests the most usable way for organizing the know-how exchange because: I) it is natural. Most of manufacturing know-how may rarely be described in a strict formalism, but it is usually shared through linguistic- or (more generally) conceptual-level representations, loosely involving the ontological information. II) It is fairly comprehensive. By the definition, an ontology can be considered as a description of the domain analysis made (or to be made) to form a given representation. The intent of an ontology is to provide the correct context for a representation, explicating the conceptualizations originally composed, restricting plausible interpretations of the representation. The ontological level is also accountable for the social context of the knowledge use that, otherwise, would be lost and that is indeed missing in many contemporary paradigms of knowledge sharing. III) It is technologically realizable. Much research in Computer Integrated Manufacturing and computer sciences is being done which is expected to promote the sharing of manufacturing know-how.

Our work offers one new contribution – clarification of the phenomenon of know-how from a computer science perspective. Our findings are based upon and supported by the literature. Hence, another contribution of this paper would be providing the reader with the introduction to the modern study of knowledge.

## 6  ACKNOWLEDGMENT

This research has been made within the International Project 'Development of the Production Design Technology for Machining' funded by NEDO, Japan. V.V. Kryssanov acknowledges the financial support of the Japan Society for the Promotion of Science under the "Methodology of Emergent Synthesis" research project (No 96P00702) in the Program "Research for the future."

## 7  REFERENCES

Berry, J.F. (1995) A Framework for Understanding Large Scale Digital Storage Systems, in *Proceedings of the 14th IEEE Symposium on Mass Storage Systems*, IEEE Computer Society Press, Los Alamitos, CA.

Cainarca, G.C., Chiesa, V. and Jovane, F. (1997) Sustainability and Production Systems: Suggestions from Empirical Cases, in *Proceedings of the 29th CIRP International Seminar on Manufacturing Systems*, pp. 33-38, Osaka University, Osaka, Japan.

Chep, A. and Anselmetti, B. (1993) A knowledge-based representation and a decision-based approach for advanced manufacturing systems, in *Proceedings of the 30th International MATADOR Conference*, pp. 535-541, UMIST, Manchester, UK.

Choo, C.W. (1995) Information Management for the Intelligent Organization: Roles and Implications for the Information Professions, in *Proceedings of the 1995 Digital Libraries Conference*, Singapore.

Frey, R.E. (1989) A philosophical framework for understanding technology. *Journal of Industrial Teacher Education*, **27** (1), 23-35.

Goguen, J. (1992) The dry and the wet, in *Proceedings of the IFIP Working Group 8.1 Conference 'Information Systems Concepts,'* (Eds. E. Falkenberg, C. Rolland and E.N. El-Sayed), pp. 1-17, Alexandria, Egypt, Elsevier North-Holland.

Gruber, T.R. (1993) A Translation Approach to Portable Ontology Specifications. *Knowledge Acquisition*, **5**, 199-220.

Guarino, N. (1994) The Ontological Level, in *Philosophy and the Cognitive Sciences*, (Eds. R. Casati, B. Smith and G. White), pp. 443-456, Hölder-Pichler-Tempsky, Vienna.

Guarino, N. (1995) Formal Ontology, Conceptual Analysis and Knowledge Representation. *International Journal of Human and Computer Studies*, *special issue on The Role of Formal Ontology in the Information Technology*, (Eds. N. Guarino and R. Poli), **43**.

Guarino, N. and Giaretta, P. (1995) Ontologies and Knowledge Bases: Towards a Terminological Clarification, in *Toward Very Large Knowledge Bases: Knowledge Building and Knowledge Sharing*, pp. 25-32, IOS Press, Amsterdam.

van Heijst, G., van der Spek, R. and Kruizinga, E. (1996) Organizing Corporate Memories, *in Proceedings of the Tenth Knowledge Acquisition Workshop (KAW'96)*, Banff, Alberta, Canada.

Herschbach, D.R. (1995) Technology as Knowledge: Implications for Instruction. *Journal of Technology Education*, **7 (1)**.

Ismail, H.S., Hon, K.K.B. and Huang, K. (1995) An Intelligent Object-Oriented Approach to the Design and Assembly of Press Tools. Annals of the CIRP, **44** (1), 91-96.

Iwata, K. and Fukuda, Y. (1986) Representation of Know-How and Its Application of Machining Reference Surface in Computer Aided Process Planning. Annals of the CIRP, **35** (1), 321-324.

Kryssanov, V.V., Abramov, V.A., Fukuda, Y. and Konishi, K. (1997) A decision-making support system based on know-how, in *Proceedings of the 29th CIRP International Seminar on Manufacturing Systems*, pp. 382-387, Osaka University, Osaka, Japan.

Kyläheiko, K. (1997) Technology management from transaction cost perspective, in *Proceedings of the 14th International Conference on Production Research*, pp. 1526-1529, IFPR, Osaka Institute of Technology & Setsunan University, Osaka, Japan.

Mahé, S., Rieu, C. and Beauchêne, D. (1996) An original model to organize know-how in a benchmarking context, in *Proceedings of the Tenth Knowledge Acquisition Workshop (KAW'96)*, Banff, Alberta, Canada.

Newell, A. (1982) The knowledge level. *Artificial intelligence*, **18** (1), 87-127.

Pirskanen, S. (1997) Practical multimedia tool for continuous development, in *Proceedings of the 14th International Conference on Production Research*, pp. 1332-1334, IFPR, Osaka Institute of Technology & Setsunan University, Osaka, Japan.

Polanyi, M. (1966) The tacit dimension, Gloucester, MA. Peter Smith.

Regli, W.C. and Gaines, D.M. (1997) National Repository for Design and Process Planning, in *Proceedings of the 1997 National Science Foundation Design & Manufacturing Grantees Conference*, Seattle, US.

Saito, K., Miyoshi, T. and Sasaki, T. (1993) Automation of Polishing Process for a Cavity Surface on Dies and Molds by Using an Expert System. Annals of the CIRP, **42** (1), 553-556.

Schulz, H. and Spath, D. (1997) Integration of Operator's Experience into NC Manufacturing. Annals of the CIRP, **46** (1), 415-418.

Sellini, F., Vargas, C. (1996) Considerations for a validation approach based on experiences of know how capitalization and of knowledge based systems (KBS) definition, in *Proceedings of the Tenth Knowledge Acquisition Workshop (KAW'96)*, Banff, Alberta, Canada.

Swartout, B., Patil, R., Knight, K. and Russ, T. (1996) Toward Distributed Use of Large-Scale Ontologies, in *Proceedings of the Tenth Knowledge Acquisition Workshop (KAW'96)*, Banff, Alberta, Canada.

Vincenti, W.G. (1984) Technological knowledge without science: The innovation of flush riveting in American airplanes, ca.1930 – ca.1950. *Technology and Culture*, **25** (3), 540-576.

Wincheringer, W. and Miklavec, M. (1996) Report-based knowledge management in decentralized production structures, in *Proceedings of the 6th IFIP TC5/ WG5.7 International Conference APMS'96*, pp. 421-424, IFIP, Kyoto, Japan.